\documentclass[table]{article}
\usepackage{spconf,amsmath,graphicx}
\usepackage[dvipsnames]{xcolor}
\usepackage{highlight2}
\usepackage{comment}
\usepackage{hyperref}
\usepackage{url}
\usepackage{enumitem}
\usepackage{booktabs}
\usepackage{mathtools}
\usepackage{dirtytalk}
\usepackage{multirow}
\usepackage[font=small]{caption}

\title{SENTENCE-SELECT: LARGE-SCALE LANGUAGE MODEL DATA SELECTION FOR RARE-WORD SPEECH RECOGNITION}
\name{W. Ronny Huang, Cal Peyser, Tara N. Sainath, Ruoming Pang, Trevor Strohman, Shankar Kumar}
\address{Google Research}

\begin{document}
%\ninept
%
\maketitle
\begin{abstract}
% \TS{i think 'long-tail' is a google thing. maybe call it rare words?}
Language model fusion helps smart assistants recognize words which are rare in acoustic data but abundant in text-only corpora (typed search logs). 
However, such corpora have properties that hinder downstream performance, including being (1) too large, (2) beset with domain-mismatched content, and (3) heavy-headed rather than heavy-tailed (excessively many duplicate search queries such as ``weather''). 
We show that three simple strategies for selecting language modeling data can dramatically improve rare-word recognition without harming overall performance. 
First, to address the heavy-headedness, we downsample the data according to a soft log function, which tunably reduces high frequency (head) sentences. 
Second, to encourage rare-word exposure, we explicitly filter for words rare in the acoustic data. 
Finally, we tackle domain-mismatch via perplexity-based contrastive selection, filtering for examples matched to the target domain. 
We down-select a large corpus of web search queries by a factor of 53x and achieve better LM perplexities than without down-selection.
When shallow-fused with a state-of-the-art, production speech engine, our LM achieves WER reductions of up to 24\% relative on rare-word sentences (without changing overall WER) compared to a baseline LM trained on the raw corpus.
These gains are further validated through favorable side-by-side evaluations on live voice search traffic.
\end{abstract}
\vspace{-5pt}
\section{Introduction}
\vspace{-5pt}
A common and frustrating failure mode of neural end-to-end ASR models is the misrecognition of words rarely seen during training. 
In order to improve rare-word performance, recent works have looked to leveraging text-only corpora as an additional source of rare-word data \cite{peyser2020improving,raju2019scalable,huang2021lookup}.
A common strategy is to incorporate these text corpora into the training of a language model which is then interpolated with the E2E model during decoding. 

\begin{figure}[!h]
  \centering
%   slides: http://shortn/_X06d2TjqhL
%   \vspace{-6pt}
  \includegraphics[width=0.99\columnwidth]{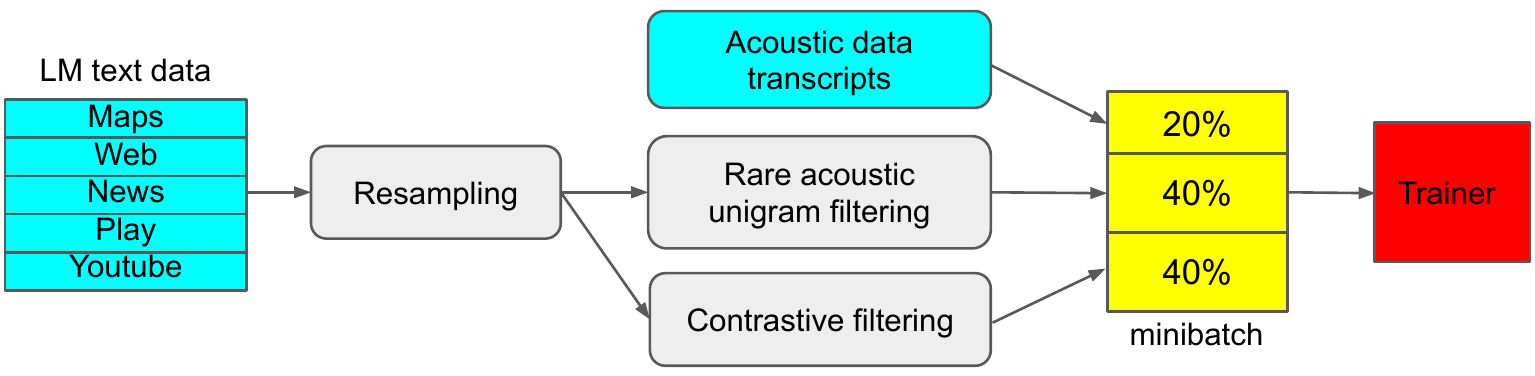}
  \vspace{-6pt}
  \caption{Overall data selection pipeline.}
  \label{fig:schem}
  \vspace{-15pt}
\end{figure}

Text data for training the language model is often scraped from typed search logs to achieve the best domain match to smart assistant voice queries. 
These logs can be very large \cite{raju2019scalable,meng2021minimum}, making it prohibitively expensive to make even a single epoch through the data, limiting rare-word exposure.
Furthermore, search queries can be heavy-headed, meaning that they contain disproportionately many high-frequency queries relative to low-frequency queries, also limiting rare-word learning.
% Such distributional bias can prevent the language model from learning the long tail of the data, including many rare words.
Finally, there is a domain mismatch between typed queries and voice queries which hurts target domain (voice) performance. 
For example, typed queries contain more website names while voice queries contain more voice commands.

In this work, we propose three simple data selection strategies applied together (Fig. \ref{fig:schem}) to reduce the size of the corpus and improve recognition quality on rare words, without hurting overall performance. 
First, we show that n\"aive deduplication \cite{lee2021deduplicating} does \textit{not} improve performance because it is too aggressive.
Instead we find beneficial a gradual downsampling of high-frequency sentences using a soft log function which maintains the natural distribution of the corpus while mitigating its heavy-headedness.
Second, we explicitly filter the dataset for sentences containing rare words in order to boost their representation during training.
Third, we use perplexity-based contrastive data selection to filter only for queries which most resemble the target domain of voice search.

% Our data selection is run on a slice of anonymized Google search traffic consisting of 213 billion sentences, or several trillion words.
% To our knowledge no other published work has performed data selection at this scale.
% % \TS{i would make sure of this statement with a literateure search. particularly look at papers from amazon, microsoft}.
% We obtain a subset that is two orders of magnitude smaller while improving results on both the head and tail of the data distribution. 
Our data selection is run on a slice of anonymized Google search traffic. We obtain a subset that is orders of magnitude smaller while improving results on both the head and tail of the data distribution.
With shallow fusion, the data-selected LM yields up to 24\% WER relative reduction on rare words compared to an LM trained on the raw corpus, while leaving the overall WER unchanged. 
Our work shows that simple data selection methods can significantly improve rare-word learning without the need for architecture or training configuration changes.
% Since data selection does not require architectural changes unlike other tail learning methods \cite{menon2020long}, it can be a drop-in replacement for existing architectures.
% \TS{i would not mention 'production' in the paper. also consider removing this sentence its not very clear what production means}.

% \input{related}
\vspace{-6pt}
\section{method}
\vspace{-6pt}
\label{sec:method}

Figure \ref{fig:schem} depicts our overall data selection pipeline.
The raw data is first downsampled by one of the functions in \S\ref{sec:downsampling} and then is used in parallel for rare acoustic unigram filtering (\S\ref{sec:filtering}) and contrastive filtering (\S\ref{sec:contrastive}).
During training, the two latter datasets, along with transcripts from the acoustic data, are combined into each minibatch according to a sampling ratio, e.g., 20\%/40\%/40\% for acoustic transcripts / rare unigram filtered / contrastive filtered. We now discuss each of the components of the pipeline.

\vspace{-8pt}
\subsection{Downsampling}
\label{sec:downsampling}
%Shankar: The colors for 'Play' (blue) and 'Maps' (green) are hard to distinguish. Can you change the color for one of these?

% We first introduce a working definition of ``rareness''
% \BL{maybe `rareness' as you are switching the term from tail to rare in the paper}
% used in the context of this paper. 
A word or sentence is more rare when it has a lower frequency (there are fewer occurrences of it) in the dataset relative to other words or sentences. 
We define ``tailedness'' to describe the relative amount of rare words or sentences in the distribution.
% Conversely it is more headed when it has a higher frequency. 
In this section on downsampling, we measure frequency at the sentence level rather than at the word level for the sake of simplicity.
In other sections, we go by word frequency.
% It is also possible to measure rareness of a sentence from an aggregate of its words; we leave this idea for future work.

\begin{figure}[t]
  \centering
%   colab: http://shortn/_ae9PlFlyfx
  \vspace{-10pt}
  \includegraphics[width=.65\linewidth]{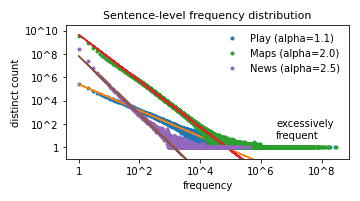} \\
  \vspace{-5pt}
  \includegraphics[width=.65\linewidth]{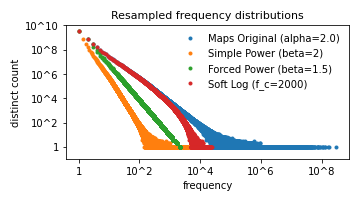}
  \vspace{-10pt}
  \caption{Top: Original frequency distributions of three data domains. Bottom: Frequency distribution of the Maps domain after downsampling with three different functions.}
  \vspace{-14pt}
  \label{fig:freqdist}
\end{figure}

We now explore the frequency distribution of our sentences to characterize the tailedness of the dataset as a whole. 
Figure \ref{fig:freqdist} (top) plots the number of distinct sentences as a function of frequency, henceforth abbreviated as $\texttt{distinct\_count}(f)$, for three different search domains. 
The frequency distribution is linear on the log-log plot, resembling a Zipfian \cite{zipf2013selected} power law, $\texttt{distinct\_count}(f) \approx Af^{-\alpha}$
% \begin{equation}
%     \label{eq:zipfian}
%     \texttt{distinct\_count}(f) \approx Af^{-\alpha},
% \end{equation}
% \noindent
where $f$ is the frequency and $A$ is the number of distinct sentences (frequency of one). 
Having a large $\alpha$ would mean a heavy-tailed distribution, since it results in fewer sentences at the high frequency part of the distribution. 
I.e. $\alpha$ approaching infinity be a fully deduplicated dataset.
Our search query datasets have $\alpha$ values between 1.1 and 2.5 depending on the domain (Fig. \ref{fig:freqdist} (top)).
For the Play domain which has an $\alpha$ of 1.1, \textit{each} of the most common sentences can occur as many times as \textit{all} of the single-frequency distinct sentences combined.
An additional observation is that the frequency distribution deviates from the Zipfian power law at very high frequencies. 
In this range, sentences tend to deviate from the power law in favor of even higher frequency. 
These \textit{excessively frequent} sentences are common search terms such as ``weather''. 
To better model tail words, we seek a downsampling function that reduces this heavy-headedness and brings out the tail while also managing these excessively common sentences.
Table \ref{tab:funcs} presents two such functions.
\vspace{3pt}

% We experiment with three downsampling functions defined in Table \ref{tab:funcs}.

\begin{table}[t]
    \small
    \caption{Downsampling functions. $f_0$ is the original frequency of a sentence and $f_1$ is the downsampled frequency.}
    \label{tab:funcs}
    \vspace{-9pt}
    \centering
    \smallskip\noindent
    % \resizebox{.99\linewidth}{!}{%
    \begin{tabular}{l|l|l}
        \toprule
        Name & Function & Param. \\
        \midrule
        Simple Power & $f_1 = f_0^{\beta}$ & $\beta$ \\
        % Forced Power & $f_1 = F(f_0)^{\beta}$  & $\beta$ \\
        Soft Log & $f_1 = f_c\log(1 + f_0 / f_c)$ & $f_c$ \\
        \bottomrule
    \end{tabular}
    % }
    \vspace{-12pt}
\end{table}

% \subsubsection{Simple Power}
\noindent
\textbf{Simple Power} tunes the tailedness of the frequency distribution via the parameter $\beta$. 
The subsequent frequency distribution will be approximately characterized by $Af^{-\alpha\beta}$.

\noindent
\textbf{Soft Log} applies logarithmic scaling after a threshold frequency $f_c$. 
We tried pure logarithmic scaling (i.e. $f_1 = \log f_0$) but found that it was too aggressive of a flattening of the distribution and reduced model performance on all metrics.
We term this function ``soft log'' because it smoothly transitions from linear scaling to logarithmic scaling.
\vspace{3pt}

In Fig. \ref{fig:freqdist} (bottom), we show example frequency distributions after downsampling (with arbitrarily picked parameters for illustration). 
Simple Power downsampling simply rescales the shape of the distribution along the x-axis, but fails to mitigate excessively frequent sentences.
% Forced Power turns the entire distribution into a perfect power law line, which involved projecting all the excessively frequent sentences onto that line.
Soft Log matches the original distribution up to around the threshold frequency $f_c$ and then cuts off the remaining (mostly excessive) frequencies.

\vspace{-8pt}
\subsection{Rare acoustic unigram filtering}
\label{sec:filtering}

\begin{figure}[b]
  \centering
%   colab: http://shortn/_UWoTnQfCuh
  \vspace{-20pt}
  \includegraphics[width=.65\linewidth]{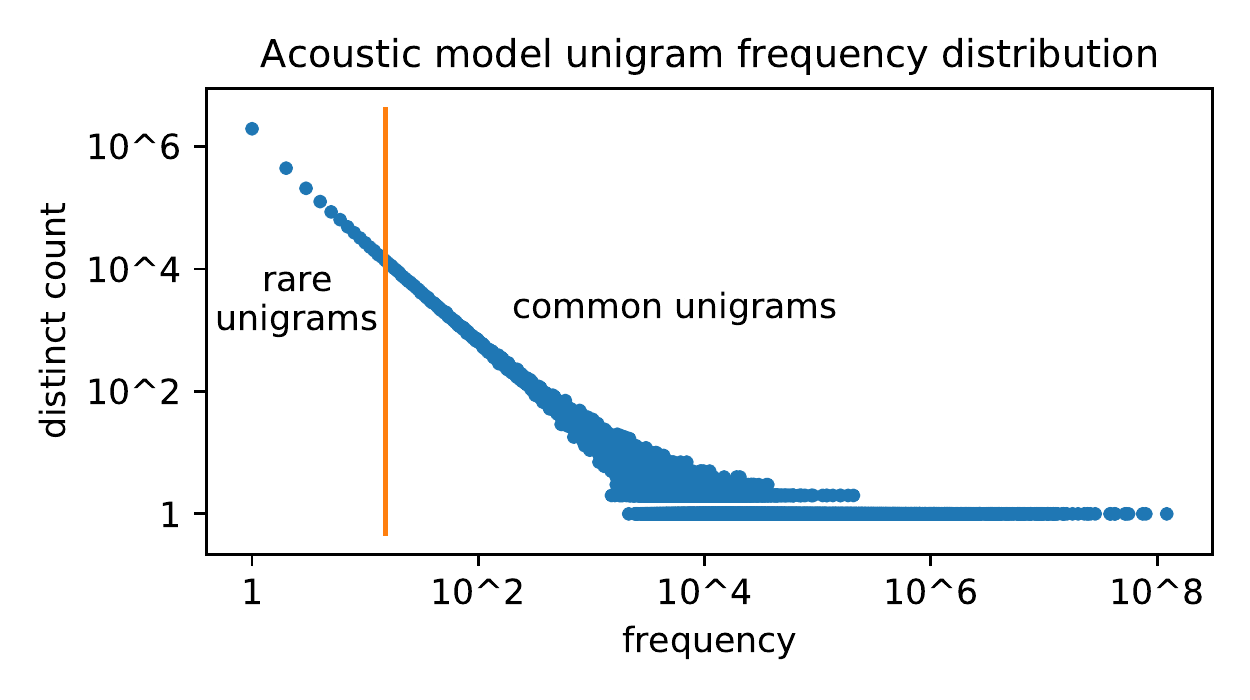}
  \vspace{-10pt}
  \caption{Unigram frequency distribution of acoustic data}
  \label{fig:unigramfreq}
  \vspace{-5pt}
\end{figure}

Since our goal is to help rare-word ASR, we can directly filter the training set for sentences which contain a unigram that is rare in the acoustic data but potentially common in the text-only data.
The acoustic model data contains 290M audio-text pairs spanning the domains of search, farfield, telephony, and YouTube \cite{narayanan2019recognizing}, and contains 3.4M distinct unigrams whose frequency distribution is shown in Fig. \ref{fig:unigramfreq}. 
We first run a spelling corrector \cite{guo2019spelling} on the text-only data in order to avoid unigrams which seem rare but are in fact misspellings.
About 3\% of all words were spell-corrected.
We then filter for a subset of the text-only data containing sentences with a unigram whose frequency in the acoustic data is less than a threshold $f_t=15$.
This threshold was tuned based on the importance of having a small resultant dataset size versus the recall of tail words.

\vspace{-10pt}
\subsection{Contrastive data filtering}
\label{sec:contrastive}

The rare-unigram-filtered sentences produced from \S\ref{sec:filtering} tend to be more domain-mismatched to the target domain of voice search, given that many such unigrams may never show up in voice search.
Further, filtering only for the tail may be a missed opportunity for extracting sentences from the corpus that may help the head.
We thus apply contrastive selection to select for sentences from the text-only corpus which are matched to the voice search domain.
In contrastive data selection \cite{Moore10,Mezzoudj18}, a score is calculated for each sentence as $\mathcal{L}_{target}(x) - \mathcal{L}_{background}(x)$.
% \vspace{-10pt}
% \begin{equation}
%     \mathrm{score}(x) = \mathcal{L}_{target}(x) - \mathcal{L}_{background}(x).
% \end{equation}
% \noindent
where $\mathcal{L}$ is the log perplexity of the sentence, $target$ and $background$ correspond to target and background LMs.
The background LM is trained on a fully deduplicated corpus to ensure that it can make several epochs through the data.
It is then fine-tuned on the acoustic model data to produce the target LM.
The score should be lower for a sentence $x$ if it is more akin to the voice domain.
Only scores below a threshold (tuned in Table \ref{tab:main}) are kept in the training data.

% We tried several minor variations of the base contrastive selection algorithm.

% \begin{itemize}
%     \item Scores are normalized by the sentence length. We divide each score by a factor of $N_{words}^\gamma$, where $\gamma$ is a number between 0 (no normalization) and 1 (full normalization). We found the optimal performance came from $\gamma=0.6$.
%     \item Sentences with fewer than $N_{words}$ words are left out. We found that $N_{words}=2$ performed slightly better than the default of $N_{words}=1$.
%     \item Perplexities from numeric tokens are prevented from contributing to the total sentence perplexity. We found that this hurt performance and left it out.
%     %Shankar: The above sentence is unclear. "We found that this hurt performance and left it out." Does this mean keeping numeric tokens hurt performance? Could you rephrase it?
% \end{itemize}
% \input{setup}
\vspace{-16pt}
\section{Results}
\label{sec:results}

% \subsection{Text corpus and data selection pipeline}
\label{sec:data}
Our raw text corpus is a slice of anonymized search traffic across the domains of Web, Maps, News, Play, and YouTube. Altogether, it consists of 213B sentences, of which 7.2B are distinct.
The data selection pipelines were written in Apache Beam and run on Cloud Dataflow \cite{dataflow2015}.
% Each part of the pipeline (gray boxes in Fig. \ref{fig:schem}) ran on 10k CPU cores for 1-4 days.

% \subsection{Model architectures}
Our E2E model, described fully in \cite{sainath2021system}, is a 150M-parameter streaming RNNT \cite{graves2012sequence} emitting 4096 lower-case wordpieces. The encoder is a cascaded \cite{narayanan2021cascaded} Conformer \cite{gulati2020conformer} and the decoder is a stateless embedding decoder \cite{botros2021tied}. 
It is trained 
% The causal encoder hypotheses are interpolated \BL{what does "interpolated" mean? the non-causal decoding does not use cause hypotheses right?}  with the hypotheses of a noncausal encoder that uses a few seconds of right context to improve quality \cite{narayanan2021cascaded}.
on 290M audio-text pairs spanning the domains of search, farfield, telephony, and YouTube \cite{narayanan2019recognizing}.
Our language model is a unidirectional 70M-parameter Conformer LM \cite{gulati2020conformer}. Both models are trained with Adam to convergence.

% \subsection{LM integration and evaluation}
\label{sec:evaluation}

We obtain an effective, LM-fused E2E score by factorizing out the speech model's log-posterior from its internal language model score and adding the LM log-posterior score as discussed in the HAT method \cite{variani2020hybrid}.
The interpolation weights ($\lambda_1$, $\lambda_2$) are optimized via blackbox optimization \cite{golovin2017google} to minimize the WERs of a VS testset and a rare proper nouns TTS testset.
Except in \S\ref{sec:sf}, LM integration is done in lattice rescoring mode rather than shallow fusion.
% For each sentence in RareASR and RareALL, we report LogPP and WER on only the rare word portion of the sentence---even though the entire sentence is passed to the model---in order to extract tail performance.
% For ASR evaluations, we synthesize TTS audio for each of the test sets. We have separately found that performance on TTS audio is correlated with recorded audio and thus believe that TTS audio is sufficient for measuring the relative performance gains of the LookupLM architecture.

\begin{table}[t]
\vspace{-1pt}
\caption{Downsampling results with a 128M parameter LM}
\vspace{-8pt}
\label{tab:downsampling}
\centering
\resizebox{\linewidth}{!}{%
\begin{tabular}{l|r|ll|ll}
\toprule
                       & Corpus          & \multicolumn{2}{c}{Log Perplexity} & \multicolumn{2}{c}{WER} \\
                       & size (B)        & VS          & TAIL        & VS        & TAIL        \\
\midrule
B0: No LM              &                                  &                              &                              & \cg{5.6}{5.3}{6.0}       & \cg{32.81}{27.25}{32.81}       \\
B1: Raw                & \cg{213.7}{7.2}{213.7}           & \cg{2.94}{2.89}{2.94}        & \cg{3.01}{2.89}{3.01}        & \cg{5.4}{5.3}{6.0}       & \cg{27.81}{27.25}{28.03}       \\
B2: Fully deduplicated & \cg{7.2}{7.2}{213.7}             & \cg{2.92}{2.89}{2.94}        & \cg{3.00}{2.89}{3.01}        & \cg{5.4}{5.3}{6.0}       & \cg{28.03}{27.25}{28.03}       \\
\midrule
E1: SimplePower-2.08   & \cg{61.6}{7.2}{213.7}            & \cg{2.93}{2.89}{2.94}        & \cg{2.93}{2.89}{3.01}        & \cg{5.4}{5.3}{6.0}       & \cg{27.31}{27.25}{28.03}       \\
E2: SimplePower-2.32   & \cg{31.9}{7.2}{213.7}            & \cg{2.93}{2.89}{2.94}        & \cg{2.92}{2.89}{3.01}        & \cg{5.4}{5.3}{6.0}       & \cg{27.32}{27.25}{28.03}       \\
E3: SimplePower-2.84   & \cg{18.4}{7.2}{213.7}            & \cg{2.90}{2.89}{2.94}        & \cg{2.90}{2.89}{3.01}        & \cg{5.4}{5.3}{6.0}       & \cg{27.39}{27.25}{28.03}       \\
% \midrule
% E4: ForcedPower-2.08   & \cg{53.6}{7.2}{213.7}            & \cg{2.92}{2.89}{2.94}        & \cg{2.91}{2.89}{3.01}        & \cg{5.4}{5.3}{6.0}       & \cg{27.29}{27.25}{28.03}       \\
% E5: ForcedPower-2.32   & \cg{32.8}{7.2}{213.7}            & \cg{2.91}{2.89}{2.94}        & \cg{2.90}{2.89}{3.01}        & \cg{5.4}{5.3}{6.0}       & \cg{27.31}{27.25}{28.03}       \\
% E6: ForcedPower-2.84   & \cg{18.9}{7.2}{213.7}            & \cg{2.89}{2.89}{2.94}        & \cg{2.90}{2.89}{3.01}        & \cg{5.4}{5.3}{6.0}       & \cg{27.40}{27.25}{28.03}       \\
\midrule
E4: SoftLog-0          & \cg{129.3}{7.2}{213.7}           & \cg{2.94}{2.89}{2.94}        & \cg{2.93}{2.89}{3.01}        & \cg{5.5}{5.3}{6.0}       & \cg{27.51}{27.25}{28.03}       \\
E5: SoftLog-0.5        & \cg{108.0}{7.2}{213.7}           & \cg{2.92}{2.89}{2.94}        & \cg{2.92}{2.89}{3.01}        & \cg{5.5}{5.3}{6.0}       & \cg{27.40}{27.25}{28.03}       \\
E6: SoftLog-1          & \cg{86.7}{7.2}{213.7}            & \cg{2.92}{2.89}{2.94}        & \cg{2.90}{2.89}{3.01}        & \cg{5.4}{5.3}{6.0}       & \cg{27.33}{27.25}{28.03}       \\
E7: SoftLog-2         & \cg{51.9}{7.2}{213.7}            & \cg{2.91}{2.89}{2.94}        & \bg{2.89}{2.89}{3.01}        & \cg{5.4}{5.3}{6.0}       & \bg{27.25}{27.25}{28.03}       \\
E8: SoftLog-3         & \cg{29.7}{7.2}{213.7}            & \bg{2.89}{2.89}{2.94}        & \bg{2.89}{2.89}{3.01}        & \cg{5.5}{5.3}{6.0}       & \cg{27.36}{27.25}{28.03}       \\
\bottomrule
\end{tabular}%
}
\vspace{-10pt}
\end{table}

\vspace{3pt}
\noindent
We report results on 2 test sets of 10k or more examples:
\begin{itemize}[leftmargin=*]
    \vspace{-3pt}
    \item VS: Voice Search, the target domain. 
    \vspace{-5pt}
    \item TAIL: TTS sentences containing a mix of difficult words, low-frequency words, or rare proper nouns in the the text data. See \S\ref{sec:sf} and \cite{peyser2020improving} for more curation details. We've found empirically that our Tacotron TTS and recorded audio have well-correlated performance. 

\vspace{-3pt}
\end{itemize}

\subsection{Downsampling results}
In Table \ref{tab:downsampling}, we evaluate log perplexity and WER on a 128M parameter LM against downsampling functions with various parameter values (appended to the function name) and their resultant corpus sizes. 
All LMs are trained on minibatches containing 50\% sentences from the text corpus (down-selected with the method of choice) and 50\% from the acoustic data transcripts. It is necessary in all of our models to include some acoustic transcripts in order to prevent VS from degrading.
For Simple Power, the parameter value shown is equal to $\alpha\beta$.
Thus a different $\beta$ is used for each domain in the corpus (Section \ref{sec:data}) since each domain has a unique $\alpha$.
For each Soft Log experiment, the parameter value shown is equal to $\log_{10}(f_r/f_c)$. $f_r$ is frequency where the power law line fit intersects 1 for each domain.
Larger parameter values indicate more aggressive mitigation of high frequency sentences.

We compare the results to three baselines. 
B0 is the standalone performance of the E2E model without an LM, showing it stands to gain from LM integration. 
B1 contains an LM trained on raw text corpus and B2 contains an LM trained on a n\"aively deduplicated text corpus.
Interestingly, in contrast to \cite{lee2021deduplicating}, deduplication does not improve performance over the raw corpus because it overly flattens the corpus's frequency distribution.

All downsampling functions match or outperform the B1 baseline in perplexity and WER, especially on TAIL.
For Simple Power, varying the parameter (tailedness) did not significantly alter the outcome.
For Soft Log, there was a sweet spot at around a parameter (log cutoff) value of 2.
This seems to suggest that the gains come primarily from mitigating excessively frequent sentences rather than from altering the tailedness of the distribution as a whole.
E7: SoftLog-2 yielded the best result (2.1\% relative WER) and a 4.1x reduction in the dataset size.
We thus used its output as the input for the remaining two components of the pipeline.

\vspace{-5pt}
\subsection{Rare unigram and contrastive filtering results}
In Table \ref{tab:main}, we evaluate the WERs for several rare-acoustic-unigram and contrastive filtered datasets and mixing ratios on a 70M parameter LM.
We compare the results against baseline B3, the performance of the downsampled dataset without additional filtering.

First, we evaluate the effect of each filtering method by itself. 
E9 shows that rare acoustic unigram filtering indeed helps improve rare word modeling, with a 6\% relative WER reduction on TAIL. It however does comes at a cost to VS (the head).
On the other hand, E10 shows that contrastive filtering for acoustic-like sentences helps VS slightly, but degrades TAIL since it likely filters out many rare words which by definition are unlike the acoustic data.
Next, we perform a contrastive selection ablation study in E11-E15.
3\% means that we keep only the top 3\% percentile of sentences based on their contrastive scores. The VS WER dependence on the threshold is small, but monotonically degrading as the threshold is loosened.
Aside from this ablation study, all contrastively selected datasets in this paper are from a threshold at the 6\% percentile which balances corpus size and performance.

Finally, E16-E18 utilize both filtered datasets, and we mix in the datasets along with the acoustic data according to different mixing ratios. 
E17 and E18 show that VS and TAIL can be simultaneously improved by our data selection methods, while E16 shows that TAIL can be significantly improved at a small cost to VS. This final corpus size is about 4B, which is 53x smaller than the original corpus.

\begin{table}[]
\vspace{-25pt}
\caption{Rare acoustic unigram and contrastive filtering results with a 70M parameter LM. Mixing ratio abbreviations: D=Downsampled, A=Acoustic, R=Rare, C=Contrastive.}
\vspace{-10pt}
\label{tab:main}
\centering
\resizebox{\linewidth}{!}{%
\begin{tabular}{l|c|c|cc}
\toprule
                         & Mixing ratio & Corpus   & \multicolumn{2}{c}{WER} \\
                         & (D, A, R, C) & size (B) & VS        & TAIL        \\
\midrule
B3: Downsampled & 50, 50, 0, 0                     & \cg{52.19}{0.5}{52.19}    & \cg{5.5}{5.3}{5.7}       & \cg{27.48}{25.87}{29.64}       \\
E9: Rare        & 0, 50, 50, 0                     & \cg{3.19}{0.5}{4}     & \cg{5.7}{5.3}{5.7}       & \cg{25.87}{25.87}{29.64}       \\
E10: Contrastive & 0, 50, 0, 50                     & \cg{0.72}{0.5}{4}     & \cg{5.3}{5.3}{5.7}       & \cg{29.52}{25.87}{29.64}       \\
\midrule
E11: Contrastive-3\%       & 0, 50, 0, 50                     & \cg{0.50}{0.5}{4}     & \cg{5.3}{5.3}{5.7}       & \cg{29.64}{25.87}{29.64}       \\
E12: Contrastive-6\%       & 0, 50, 0, 50                     & \cg{0.72}{0.5}{4}     & \cg{5.3}{5.3}{5.7}       & \cg{29.52}{25.87}{29.64}       \\
E13: Contrastive-9\%       & 0, 50, 0, 50                     & \cg{0.93}{0.5}{4}     & \cg{5.3}{5.3}{5.7}       & \cg{29.41}{25.87}{29.64}       \\
E14: Contrastive-16\%      & 0, 50, 0, 50                     & \cg{1.43}{0.5}{4}     & \cg{5.4}{5.3}{5.7}       & \cg{29.38}{25.87}{29.64}       \\
E15: Contrastive-30\%      & 0, 50, 0, 50                     & \cg{2.42}{0.5}{4}     & \cg{5.4}{5.3}{5.7}       & \cg{29.39}{25.87}{29.64}       \\
\midrule
E16: Mix-20/40/40          & 0, 20, 40, 40                    & \cg{3.91}{0.5}{4}     & \cg{5.6}{5.3}{5.7}       & \cg{26.26}{25.87}{29.64}       \\
E17: Mix-40/20/40          & 0, 40, 20, 40                    & \cg{3.91}{0.5}{4}     & \cg{5.4}{5.3}{5.7}       & \cg{27.03}{25.87}{29.64}       \\
E18: Mix-40/40/20          & 0, 40, 40, 20                    & \cg{3.91}{0.5}{4}     & \cg{5.4}{5.3}{5.7}       & \cg{26.77}{25.87}{29.64} \\
\bottomrule
\end{tabular}%
}
\vspace{-12pt}
\end{table}

\vspace{-5pt}
\subsection{Rescoring vs. shallow fusion}
\label{sec:sf}
The results up to now have used LM-based lattice rescoring as the LM fusion method.
While rescoring is more computationally efficient, it prevents the beam search from utilizing the full potential of the LM.
For example, rare words that appear at the beginning or middle of the sentence may be pruned from the beam before the LM has the chance to score it at the end.
Shallow fusion better utilizes the LM since it uses the LM to score at each token.

Table \ref{tab:rescsf} compares rescoring and shallow fusion for the E16 and E17 models from Table \ref{tab:main}, as well as a for a baseline trained on the downsampled data.
To better understand where our gains are, we break TAIL into its four constituent datasets, each containing a different category of rare word:
acoustic-rare and text-only-common (TMC), acoustic-rare and text-only-rare (TMR), rare proper nouns (RPN) as determined by a tagger, and surprising pronunciations (SP).

Overall, shallow fusion achieves significantly better results on TAIL than rescoring as expected.
The filtered datasets (E17-18) with shallow fusion outperform the unfiltered baseline (B3) on all TAIL sets except for RPN, which may be due to our rare word filtering having too low of a recall compared to the rare proper noun tagger.
For the favorable categories, E17 achieves between 7\%-24\% relative improvement with the best performance, 24\% relative, being on the TMC dataset which contains words which are rare in the acoustic data but common in the text-only data.
E17 also leaves the head, VS, unchanged.

\begin{table}[]
\vspace{-25pt}
\caption{Rescoring vs. Shallow fusion.}
\vspace{-10pt}
\label{tab:rescsf}
\centering
\resizebox{\linewidth}{!}{%
\begin{tabular}{l|l|ccccc}
\toprule
                              \multicolumn{2}{r}{Fusion}  & \multicolumn{5}{c}{WER}             \\
                              \multicolumn{2}{r}{method}  & VS  & TMC   & TMR   & RPN   & SP    \\
\midrule
\multirow{2}{*}{B3: Downsampled} & Resc.               & \cg{5.5}{5.3}{5.7} & \cg{27.8}{19.6}{27.8}  & \cg{46.3}{40.1}{46.3} & \cg{16.6}{15.9}{17.6} & \cg{19.2}{15.8}{19.2} \\
                              & SF                & \cg{5.6}{5.3}{5.7} & \cg{25.8}{19.6}{27.8}  & \cg{45.8}{40.1}{46.3} & \bg{15.9}{15.9}{17.6} & \cg{18.1}{15.8}{19.2} \\
\midrule
\multirow{2}{*}{E17: Mix-20/40/40} & Resc.           & \cg{5.6}{5.3}{5.7} & \cg{25.6}{19.6}{26.5}  & \cg{43.6}{40.1}{45.2} & \cg{17.4}{15.9}{17.6} & \cg{18.3}{15.8}{19.2} \\
                              & SF                & \cg{5.6}{5.3}{5.7} & \bg{19.6}{19.6}{26.5}  & \bg{40.1}{40.1}{45.2} & \cg{16.2}{15.9}{17.6} & \bg{15.8}{15.8}{19.2} \\
\midrule
\multirow{2}{*}{E18: Mix-40/20/40} & Resc.           & \bg{5.4}{5.3}{5.7} & \cg{26.5}{19.6}{26.5}  & \cg{45.2}{40.1}{45.2} & \cg{17.6}{15.9}{17.6} & \cg{18.7}{15.8}{19.2} \\
                              & SF                & \cg{5.5}{5.3}{5.7} & \cg{21.4}{19.6}{26.5}  & \cg{42.2}{40.1}{45.2} & \cg{16.5}{15.9}{17.6} & \cg{16.4}{15.8}{19.2} \\
\bottomrule
\end{tabular}%
}
\vspace{-15pt}
\end{table}

\vspace{-6pt}
\subsection{Side-by-side evaluation}
% To better compare performance of our proposed CascEnc+LM model to conventional, we perform a ``side-by-side'' (SxS) on previously unseen utterances.
To measure whether the filtering methods have a positive impact on real traffic, we perform a side-by-side evaluation in which we collect 500 utterances where B3 and E17 outputs differ.
Each transcription is rated by two humans as a win (i.e. E17 is correct and B3 is not), loss (vice-versa), or neutral (both or neither are correct).

There is a small, modestly significant enhancement on the overall live traffic.
The transcriptions differed in 7.8\% of the traffic. Of the 500 collected diffs, E17 won 32 times and lost 20 times versus to the baseline, for a p-value of 0.2-0.3.
We inspect individual transcripts in Table \ref{tab:decodes} to assess the win/loss modalities.
The first two losses (left column) are due to the LM being overly eager to emit rare words. The third loss occurs at the first word where the LM typically has the highest perplexity (it is difficult to predict the first word without any context), indicating that the issue is due to the LM architecture (unidirection rather than bidirectional) instead of the data.
The wins from data selection tend to be on rare words as shown on the right column.

% \begin{table}[h]
%     \caption{Side-by-side results.}
%     \label{tab:sxs}
%     \centering
%     \smallskip\noindent
%     \resizebox{.6\linewidth}{!}{%
%     \begin{tabular}{llll}
%         \toprule
%         Changed & Wins & Losses & p-value \\
%         \midrule
%         7.8\% & 32 & 20 & 0.2-0.3 \\ 
%         \bottomrule
%     \end{tabular}}
%     \label{tab:decodes}
% \end{table}

\begin{table}[h]
    \vspace{-4pt}
    \caption{Decoding samples from side-by-side.}
    \vspace{-12pt}
    \label{tab:decodes}
    \centering
    \smallskip\noindent
    \resizebox{.8\linewidth}{!}{%
    \begin{tabular}{ll|ll}
        \toprule
        \multicolumn{2}{c}{Data selection losses} & \multicolumn{2}{c}{Data selection wins} \\
        Baseline & Selected & Baseline & Selected \\
        \midrule
        \begin{tabular}[c]{@{}l@{}}call taco\\\textcolor{ForestGreen}{la bala}\end{tabular} & \begin{tabular}[c]{@{}l@{}}call taco\\\textcolor{red}{labella}\end{tabular} & \begin{tabular}[c]{@{}l@{}}turn on\\\textcolor{red}{chromecast} tv\end{tabular} & \begin{tabular}[c]{@{}l@{}}turn on\\\textcolor{ForestGreen}{konigsberg} tv\end{tabular} \\
        \midrule
        \begin{tabular}[c]{@{}l@{}}lake tahoe\\\textcolor{ForestGreen}{live cam}\end{tabular} & \begin{tabular}[c]{@{}l@{}}lake tahoe\\\textcolor{red}{lifetime}\end{tabular} & \begin{tabular}[c]{@{}l@{}}what is a\\\textcolor{red}{monkey}\end{tabular} & \begin{tabular}[c]{@{}l@{}}what is a\\\textcolor{ForestGreen}{makikoshi}\end{tabular}  \\
        \midrule
        \begin{tabular}[c]{@{}l@{}}\textcolor{ForestGreen}{bbc}\\ radio 2\end{tabular} & \begin{tabular}[c]{@{}l@{}}\textcolor{red}{nbc}\\ radio 2\end{tabular} & \begin{tabular}[c]{@{}l@{}}donald\\\textcolor{red}{trump}\end{tabular} & \begin{tabular}[c]{@{}l@{}}donald\\\textcolor{ForestGreen}{fredland}\end{tabular}  \\
        \bottomrule
    \end{tabular}}
    \vspace{-10pt}
\end{table}
\vspace{-10pt}
\section{Conclusion}
\vspace{-5pt}
Using scalable LM data selection methods, we reduce the size of a corpus by 53x with no degradation in overall WER while improving rare word recognition by up to 24\% relative compared to no filtering.
% Possible future directions include applying the Soft Log resampling strategy at the word level rather than sentence level, or applying the rare-unigram and contrastive filtering in series rather than in parallel to further reduce the size of the dataset.

\vfill\pagebreak

\bibliographystyle{IEEEbib}
\bibliography{refs}

\end{document}